\documentclass[journal,11pt]{IEEEtran}

\usepackage{amsmath,amssymb}
\usepackage{graphicx}
\usepackage{cite}
\usepackage{xcolor}
\usepackage{soul}
\usepackage{enumitem}
\usepackage{tabularx}
\usepackage{array}
\usepackage[hidelinks,breaklinks=true]{hyperref}

\title{FLAT: Revealing Hidden Latent-Conditioned \\ Backdoor Failures in Federated Learning}

\author{Tuan Nguyen,
        Sze Jue Yang,
        Khoa D. Doan,
        Chee Seng Chan,
        and Kok-Seng Wong%
\thanks{Corresponding authors: Chee Seng Chan and Kok-Seng Wong.}
\thanks{Tuan Nguyen is with the College of Engineering and Computer Science, VinUniversity, Hanoi, Vietnam; the VinUni-Illinois Smart Health Center, VinUniversity, Hanoi, Vietnam; and the Siebel School of Computing and Data Science, University of Illinois Urbana-Champaign, Urbana, IL, USA.}
\thanks{Sze Jue Yang is with the Faculty of Computer Science and Information Technology, Universiti Malaya, Kuala Lumpur, Malaysia.}
\thanks{Chee Seng Chan is with the Faculty of Computer Science and Information Technology, Universiti Malaya, Kuala Lumpur, Malaysia, and the College of Engineering and Computer Science, VinUniversity, Hanoi, Vietnam. Email: cs.chan@um.edu.my.}
\thanks{Khoa D. Doan and Kok-Seng Wong are with the College of Engineering and Computer Science, VinUniversity, Hanoi, Vietnam, and the VinUni-Illinois Smart Health Center, VinUniversity, Hanoi, Vietnam. Email: wong.ks@vinuni.edu.vn.}
\thanks{Code is available at \url{https://github.com/sail-research/flat}.}
}

\begin{document}

\maketitle

\begin{abstract}
Horizontal federated learning (HFL) backdoor audits often summarize model behavior through clean accuracy (CA), mean attack success rate (ASR), or a single known-trigger test. Such summaries can hide a different failure mode, in which one target label is activated by many trigger realizations. We study this failure mode with FLAT, a latent-conditioned reliability stress test for HFL backdoors. In FLAT, compromised clients still submit ordinary classifier updates to the server, while an attacker-side generator $G(x,t,z)$ separates target intent $t$ from trigger realization $z$. This separation shifts the audit question from whether one known trigger succeeds to how the hidden behavior varies across targets, latent samples, defenses, and post-stop rounds. On CIFAR-10, CIFAR-100, and Tiny-ImageNet, FLAT preserves clean utility while reaching 99.49\%, 99.66\%, and 94.10\% single-target FedAvg ASR. The evaluation also reveals non-uniform defense responses, where a server rule can suppress one target mode while leaving another active. These observations motivate HFL backdoor audits that report target-wise ASR, worst-target ASR, target coverage, latent-sampled behavior, post-stop persistence, and defense response.
\end{abstract}


\section{Introduction}
In HFL, the server trains a shared model from client updates while the training samples remain on the clients. Although this design reduces direct data exposure, it also makes backdoor behavior difficult to audit. The server may observe aggregate validation behavior, update norms, and participation records, but it typically cannot inspect the local data or the exact local objective used by each client. As a result, a compromised global model can pass CA checks while still failing on inputs that contain an attacker-chosen trigger.

Current HFL backdoor auditing often relies on CA, mean ASR, or single-trigger tests. These checks can miss a broader hidden failure mode. That is, the same attacker-chosen target label may be activated by multiple trigger realizations, so the model hosts not a single fixed patch-to-label rule, but a distribution of target-conditioned behaviors. Prior federated learning (FL) backdoor studies show why this audit problem is becoming broader. A malicious client can poison local data or scale its model update~\cite{bagdasaryan2020backdoorfl}; several malicious clients can split a trigger across participants~\cite{xie2020dba}; and later attacks can aim for persistence or adaptation after the attack window~\cite{zhang2022neurotoxin,nguyen2023iba,zhang2023a3fl}. Arbitrary-target and target-on-demand work further shows that the adversary need not be limited to one fixed target label~\cite{doan2022marksman,nguyen2024venomancer}.

\begin{figure*}[t]
\centering
\includegraphics[width=0.95\textwidth, keepaspectratio]{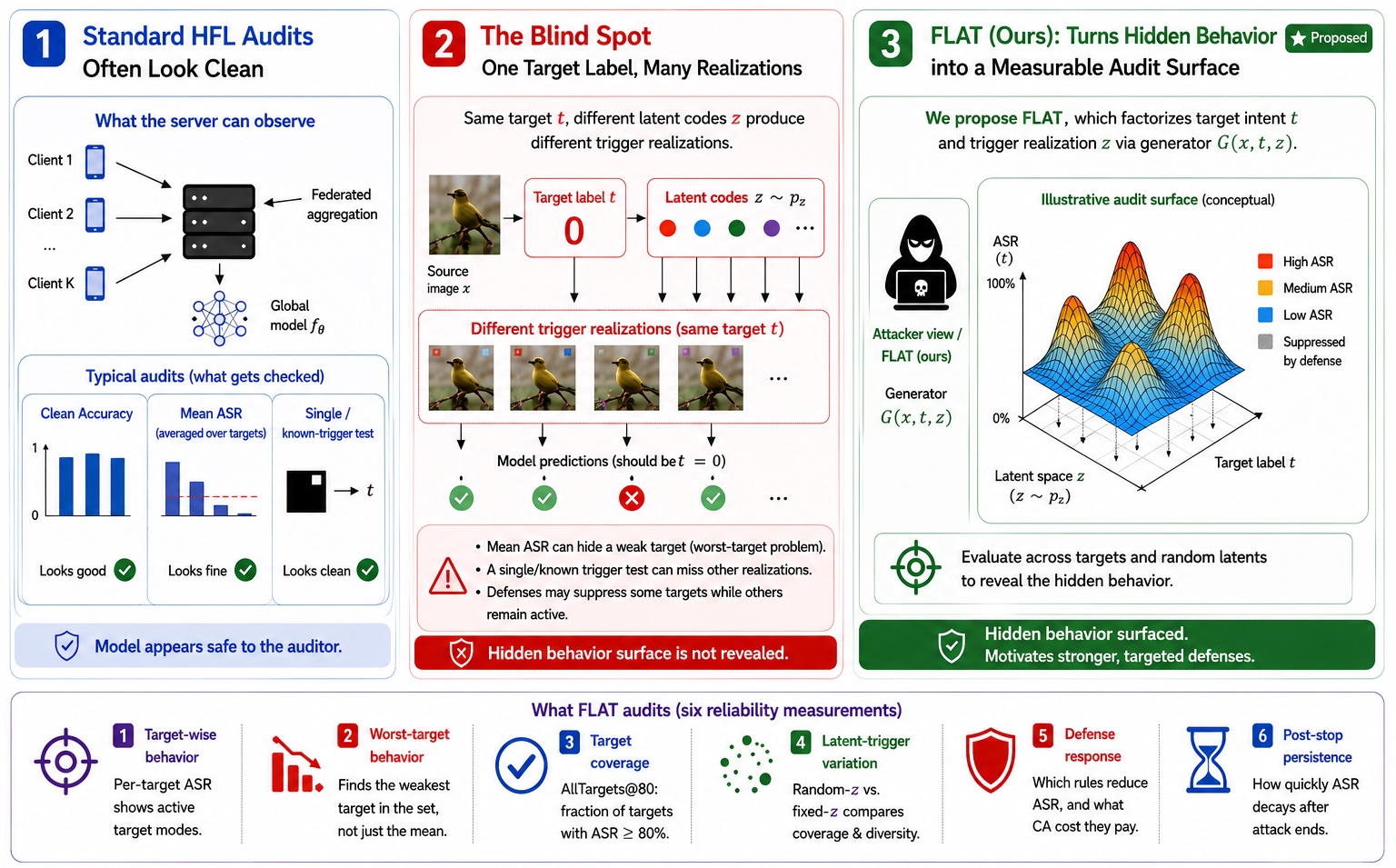}
\caption{Conceptual overview of FLAT. Standard HFL audits may check CA, mean ASR, or one known trigger and still miss a target-conditioned behavior surface. FLAT separates target intent $t$ from latent realization $z$ through $G(x,t,z)$ and audits that surface through target-wise behavior, worst-target behavior, target coverage, latent-trigger variation, defense response, and post-stop persistence.}
\label{fig:teaser}
\end{figure*}

Within that broader family, our focus is narrower. Existing target-conditioned attacks mainly test whether the attacker can choose the target label, whereas FLAT asks what happens when target intent and trigger realization are separated. For a source input $x$, target label $t$, and latent code $z$, an attacker-side generator $G(x,t,z)$ produces one realization from a family of target-conditioned perturbations. The target label specifies the intended misclassification, while the latent code selects how that target behavior is realized. The resulting factorization turns latent-conditioned target behavior into a measurable audit surface rather than a single trigger instance. Fig.~\ref{fig:teaser} summarizes this audit blind spot and how FLAT exposes it.

Once target intent and trigger realization are separated, the measurement problem changes as well. Mean ASR can hide a weak target, and a single trigger sample can hide latent modes that remain active. A defense may also suppress one part of the behavior surface while leaving another part exposed. We therefore evaluate FLAT with target-wise and worst-target ASR, target coverage, latent-sampled behavior through random-$z$ versus fixed-$z$ comparisons, post-attack persistence, and defense response. Experiments on CIFAR-10, CIFAR-100, and Tiny-ImageNet show that FLAT can achieve high attack success while preserving clean utility. More importantly, defenses do not act uniformly across targets. These findings motivate an audit that goes beyond aggregate checks.

We make the following three contributions.
\begin{itemize}
\item We identify a blind spot in current HFL backdoor auditing, where CA, mean ASR, and single-trigger tests can miss target-conditioned behavior induced by multiple trigger realizations for the same label.
\item We instantiate this risk with FLAT, a generator-based stress-test attack that factorizes target intent and trigger realization as $G(x,t,z)$, enabling latent-conditioned behavior to be evaluated systematically.
\item We show why this audit is necessary. FLAT achieves high attack success across CIFAR-10, CIFAR-100, and Tiny-ImageNet, and defense-response analysis reveals non-uniform failure surfaces, where some target modes are suppressed while others remain active under the same defense.
\end{itemize}

Section~\ref{sec:background} reviews FL backdoors, target-on-demand attacks, and defenses. Section~\ref{sec:threat-model} gives the threat model and problem formulation. Section~\ref{sec:flat-method} describes FLAT. Section~\ref{sec:experimental-setup} defines the experimental setup. Section~\ref{sec:results} reports the empirical analysis. Section~\ref{sec:discussion} discusses implications and limitations, and Section~\ref{sec:conclusion} concludes.

\section{Background and Related Work}
\label{sec:background}
\label{background}

\subsection{Backdoor Attacks in HFL}
HFL gives the server a narrow audit view, since it observes selected-client metadata, returned updates, and aggregate validation behavior rather than the local data or the local training objective. FedAvg~\cite{mcmahan2017fedavg} made this client-update protocol a practical training rule, and FedProx~\cite{li2020fedprox} and SCAFFOLD~\cite{karimireddy2020scaffold} later studied how client heterogeneity affects optimization and convergence. In this setting, a backdoor is an assurance problem as well as an attack mechanism because the global model may retain high CA while responding to a hidden condition chosen by the adversary.

Centralized backdoor studies established fixed, learned, and input-aware trigger patterns, including BadNets~\cite{gu2017badnets}, Trojaning~\cite{liu2018trojannn}, and input-aware dynamic triggers~\cite{nguyen2020inputaware}. HFL changes the threat surface because malicious behavior enters through client training and aggregation. Model-replacement and update-scaling attacks~\cite{bagdasaryan2020backdoorfl} showed that poisoned local training can survive aggregation, DBA~\cite{xie2020dba} split trigger behavior across clients, and Attack of the Tails~\cite{wang2020attackoftails} showed that rare but meaningful subpopulations can fail even when average validation behavior is benign. Together, these FL attacks illustrate the core risk that a server-visible update can look useful for clean validation while carrying behavior that appears only under a hidden condition.

Persistence, stealth, and adaptation add a second axis to this risk. Neurotoxin~\cite{zhang2022neurotoxin} masks high-clean-gradient coordinates to reduce overwriting, Chameleon~\cite{dai2023chameleon} adapts triggers to peer images for durability, and IBA~\cite{nguyen2023iba} studies backdoors that remain difficult to erase after the attack window. A3FL~\cite{zhang2023a3fl} and gradient-alignment attacks~\cite{yang2025lga} examine adaptive behavior under defense pressure and update statistics, while Cerberus~\cite{lyu2023cerberus}, FCBA~\cite{liu2024fcba}, BATMAN~\cite{he2026batman}, and Less is More~\cite{ye2026lessismore} study collusion, self-adaptation, or persistence under different FL assumptions. This literature provides the HFL backdoor context for FLAT, whose focus is the less directly measured case in which a reusable attacker-side generator and a latent input make the target behavior distributional.

\subsection{Target-Conditional and Dynamic Trigger Backdoors}
Early backdoor evaluations often fix both the target label and the trigger pattern, which gives a clean measurement protocol but can understate the behavior exposed when the adversary chooses a target label or varies the trigger realization. Marksman~\cite{doan2022marksman} established arbitrary-target conditioning in backdoor attacks, and Venomancer~\cite{nguyen2024venomancer} brought target-on-demand imperceptible trigger generation into the FL setting. These works are closest to FLAT along the target-choice axis, while input-aware dynamic triggers~\cite{nguyen2020inputaware} and Chameleon~\cite{dai2023chameleon} are closest along the trigger-realization axis.

FLAT connects those axes by treating target intent and trigger realization as separate variables. Prior target-conditioned methods primarily evaluate whether an attacker can produce a successful triggered input for a requested target $t$. FLAT evaluates the additional behavior induced when the same source image $x$ and target $t$ are paired with different latent samples $z$, so that $G(x,t,z)$ defines a set of possible trigger realizations for the same requested failure. Under this view, one sampled trigger for one target is only a partial audit of the hidden behavior.

Multi-target and collusive FL attacks further motivate target-wise measurement. DBA~\cite{xie2020dba} splits trigger behavior across clients, Cerberus~\cite{lyu2023cerberus} and CoBA~\cite{lyu2025coba} study coordinated malicious clients, Mirage~\cite{li2025mirage} studies multi-label conflict among FL backdoors, and SADBA~\cite{feng2025sadba} adds self-adaptive distributed trigger behavior. Against this literature, FLAT isolates generator reuse and latent-conditioned trigger variation within the paper's HFL operating points, where target-wise behavior, latent sampling, and post-stop persistence are evaluated under matched protocol choices.

\subsection{Federated Defenses and Evaluation Metrics}
Server-side defenses are common in FL because the server cannot inspect private client data. Robust aggregation rules such as coordinate median~\cite{yin2018byzantine}, trimmed mean~\cite{yin2018byzantine}, Krum~\cite{blanchard2017krum}, Multi-Krum~\cite{blanchard2017krum}, and RFA~\cite{pillutla2022rfa} reduce the influence of outlying or geometrically isolated updates. Similarity- and trust-based defenses such as FoolsGold~\cite{fung2020foolsgold} and FLTrust~\cite{cao2021fltrust} use historical update similarity or trusted server data to adjust client influence, while clipping and noise mechanisms~\cite{abadi2016dp}, certification-oriented methods such as CRFL~\cite{xie2021crfl}, and robust learning-rate rules~\cite{ozdayi2021robustlr} restrict how much a malicious update can affect the global model. Backdoor-specific defenses add clustering, validation, direction checks, representation analysis, or model-inspection signals, as in FLAME~\cite{nguyen2022flame}, DeepSight~\cite{rieger2022deepsight}, FLIP~\cite{zhang2023flip}, Lockdown~\cite{huang2023lockdown}, BackdoorIndicator~\cite{li2024backdoorindicator}, FLShield~\cite{kabir2024flshield}, AlignIns~\cite{xu2025alignins}, MARS~\cite{wan2025mars}, and G2uardFL~\cite{yu2026guardfl}.

These defenses motivate evaluation beyond CA and a single aggregate ASR. In the latent-conditioned setting studied here, CA does not evaluate triggered behavior, average ASR can hide a weak target, and one latent draw can miss another realization that activates the same target. FLAT reports target-wise ASR, mean ASR, worst-target ASR, target coverage, latent ablations, post-stop ASR, and defense response. The metric set is not intended as a universal defense benchmark, but as an audit surface for trained HFL image classifiers whose hidden behavior may vary across targets, latent samples, server rules, and post-stop rounds.

\section{Threat Model and Problem Formulation}
\label{sec:threat-model}
\subsection{HFL Setting}
We study HFL for $C$-class image classification. A server coordinates $K$ clients but does not read their training samples. Client $k$ holds $\mathcal{D}_k=\{(x_i,y_i)\}_{i=1}^{n_k}$, drawn from a local distribution that may differ from other clients' distributions. In round $r$, the server sends the current global model $f_{\theta^r}$ to selected clients $\mathcal{S}^r$. Let $\theta_k^{r,\mathrm{loc}}$ be the local model obtained by selected client $k$ after training from $\theta^r$; the returned update is $\Delta_k^r=\theta_k^{r,\mathrm{loc}}-\theta^r$. The server aggregates the returned updates as
\begin{equation}
    \theta^{r+1}
    =
    \mathcal{A}\left(\theta^r,\{\Delta_k^r:k\in\mathcal{S}^r\}\right).
\end{equation}
FedAvg is the reference aggregation rule in standard FL and in many FL backdoor studies~\cite{mcmahan2017fedavg,bagdasaryan2020backdoorfl}. In defense experiments, $\mathcal{A}$ may be replaced with a specified aggregation, clipping, noise, or filtering rule.

For a clean test set $\mathcal{D}_{\mathrm{test}}$, clean utility is measured by
\begin{equation}
    \mathrm{CA}(f_\theta)
    =
    \Pr_{(x,y)\sim\mathcal{D}_{\mathrm{test}}}
    \left[\arg\max_c f_\theta(x)_c = y\right].
\end{equation}

FLAT tests a failure mode missed by CA, in which the model classifies clean inputs correctly while an attacker-chosen transform $T(x,t,z)$ drives the prediction toward target label $t$.

\subsection{Attacker Goals, Capabilities, and Constraints}
The adversary controls a small subset of clients $\mathcal{M}\subset\{1,\ldots,K\}$ and participates only through the normal FL client interface. In round $r$, the selected malicious clients are $\mathcal{M}^r=\mathcal{M}\cap\mathcal{S}^r$. The server is assumed to follow the FL protocol and aggregation rule, but it does not know which selected clients are malicious.

The attacker receives the broadcast global model and knows the label space, training schedule, target policy, and the aggregation or defense rule being evaluated. It observes only its own local data. If a defense is hidden from the attacker, that case is reported separately as a black-box evaluation. FLAT is a local data/model poisoning attack: a selected malicious client can create triggered local examples, change its local loss, and submit a shaped update. It cannot edit benign clients' data, change the server code, or inspect private data from uncompromised clients. We also do not assume update scaling or model replacement unless that capability is named as an experimental variant. These assumptions match the FL backdoor setting used by fixed-trigger, distributed, durable, and adaptive attacks~\cite{bagdasaryan2020backdoorfl,xie2020dba,zhang2022neurotoxin,nguyen2023iba,zhang2023a3fl}.

Let $\mathcal{Y}=\{0,\ldots,C-1\}$ and let $\mathcal{T}\subseteq\mathcal{Y}$ denote the predefined target set; in an all-class evaluation, $\mathcal{T}=\mathcal{Y}$. We adopt the arbitrary-target setting: after training, the adversary may choose $t\in\mathcal{T}$ without retraining the global model or the trigger generator. For a clean input $x$ whose true label is not $t$, the adversary applies $T(x,t,z)$, where $z\in\mathcal{Z}$ is sampled from $p_z$. The dimension and distribution family of $p_z$ are set in the experimental protocol. The attack succeeds on $(x,t,z)$ when
\begin{equation}
    \arg\max_c f_\theta(T(x,t,z))_c = t.
\end{equation}
A fixed-trigger backdoor uses one trigger pattern, while target-conditional attacks allow the adversary to specify $t$. FLAT keeps target choice and trigger realization as separate variables. $t$ names the intended class, while $z$ indexes one trigger realization for the same input-target pair.

\subsection{Latent-Conditioned Reliability Failure}
Let $G_\phi$ denote a trigger generator with parameters $\phi$. Given a source input $x$, target label $t$, and latent code $z$, FLAT defines
\begin{align}
    \delta_\phi(x,t,z)\hphantom{,x)}
    &= G_\phi(x,t,z),\\
    T(x,t,z)\hphantom{d(,x)}
    &= \Pi_{\mathcal{X}}\left(x+\delta_\phi(x,t,z)\right),\\
    d(T(x,t,z),x)
    &\le \epsilon,
\end{align}
where $\Pi_{\mathcal{X}}$ projects the triggered input back to the valid input domain and $d(\cdot,\cdot)$ is the perturbation or perceptual distance used in the experiment. For norm-bounded variants, $d$ may be instantiated as an $\ell_\infty$ or $\ell_2$ constraint. The label $t$ specifies the attacker's intended class. The latent code $z$ selects one perturbation from the generator. For each target $t$, $G_\phi$ therefore defines a distribution over triggered inputs instead of one deterministic trigger. Section~\ref{sec:flat-method} instantiates this abstract map by using a normalized generator $\tilde{G}_\phi(x,e(t),z)$ with an explicit target-label embedding $e(t)$ and scaling its output by the perturbation budget $\epsilon$.

\begin{figure*}[t]
\centering
\includegraphics[width=0.94\textwidth, keepaspectratio]{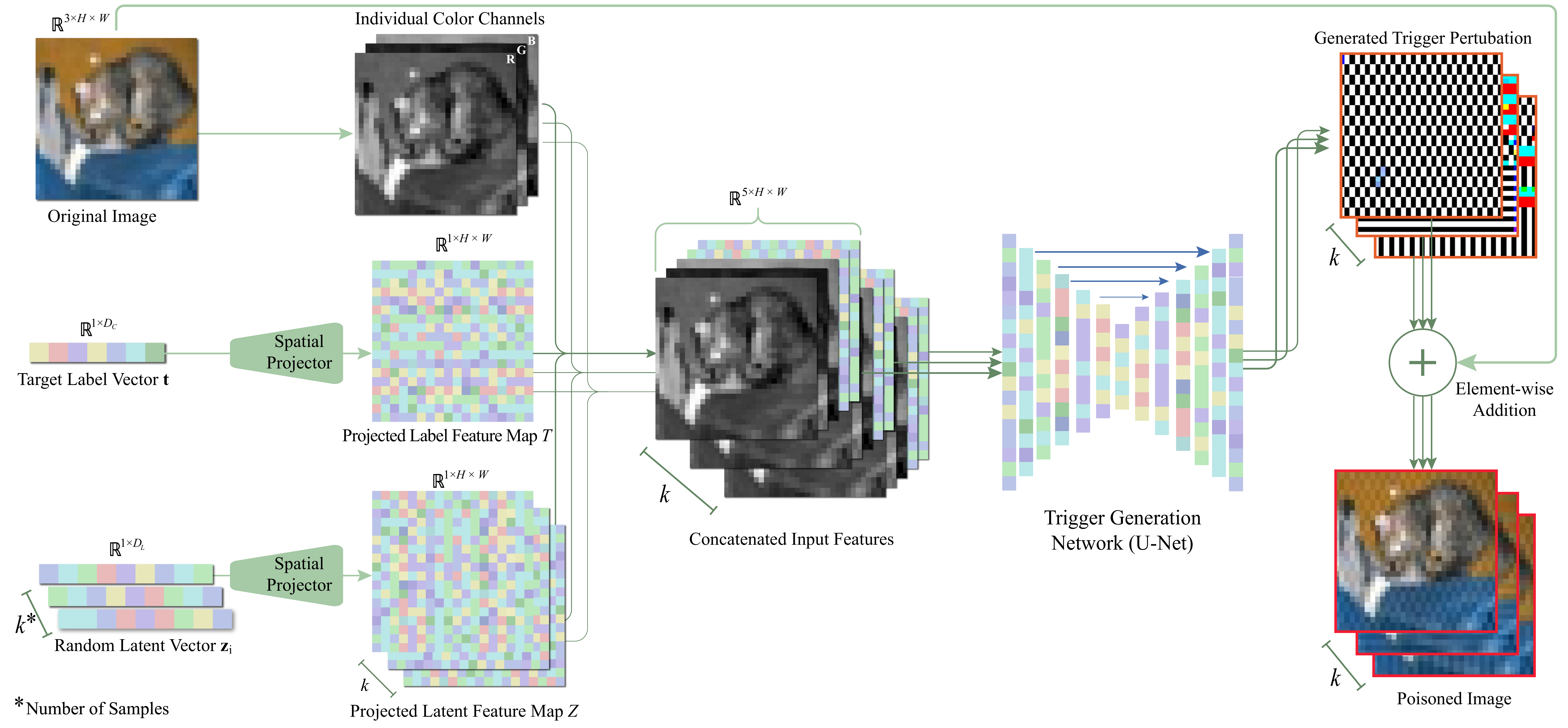}
\caption{Latent-conditioned trigger generator in FLAT. A source image, target-label representation, and latent code are projected into spatial feature maps, concatenated, and passed through a generator that outputs a bounded perturbation. The triggered (poisoned) image is formed by adding this perturbation to the original image and projecting back to the valid input domain.}
\label{fig:flat-overview}
\end{figure*}

For compactness, let $h_\theta(u)=\arg\max_c f_\theta(u)_c$. For a fixed target $t$, the target-specific ASR is
\begin{equation}
    \mathrm{ASR}(t)
    =
    \Pr_{\substack{(x,y)\sim\mathcal{D}_{\mathrm{test}},\,y\neq t\\
    z\sim p_z}}
    \left[h_\theta(T(x,t,z))=t\right].
\end{equation}
Mean ASR is not enough when $\mathcal{T}$ contains several labels. A high average can hide a failed target, a target that works only for a narrow subset of latent codes, or a trigger family that is easy to detect for some targets. We therefore report both target-averaged and worst-target success rates:
\begin{align}
    \mathrm{ASR}_{\mathrm{avg}}
    &=
    \frac{1}{|\mathcal{T}|}\sum_{t\in\mathcal{T}}\mathrm{ASR}(t),\\
    \mathrm{ASR}_{\mathrm{worst}}
    &=
    \min_{t\in\mathcal{T}}\mathrm{ASR}(t).
\end{align}

For persistence, let $r_{\mathrm{stop}}$ be the last attack round and let $\tau$ denote the number of subsequent benign rounds. With $\theta_\tau=\theta^{r_{\mathrm{stop}}+\tau}$, we measure
\begin{equation}
    \mathrm{ASR}_{\mathrm{stop}}(t,\tau)
    =
    \Pr_{\substack{(x,y)\sim\mathcal{D}_{\mathrm{test}},\,y\neq t\\
    z\sim p_z}}
    \left[h_{\theta_\tau}(T(x,t,z))=t\right].
\end{equation}

\section{FLAT: Latent-Driven Arbitrary-Target Backdoor Attack}
\label{sec:flat-method}
\subsection{Overview}
Each selected malicious client trains a local classifier copy while using an attacker-maintained trigger generator shared across compromised clients. Because the generator evolves during training, this section writes the triggered transform as $T_\phi$, and after training, once $\phi$ is fixed, drops the subscript and writes $T$. In round $r$, client $k$ starts from the broadcast model $f_{\theta^r}$ and generator state $\phi^r$, adapts the generator locally to obtain $\hat{\phi}_k^r$, and then constructs mixed clean and poisoned batches using $T_{\hat{\phi}_k^r}(x,t,z)$. The client trains its local classifier on those batches and submits only the classifier update $\Delta_k^r$ to the server. Figure~\ref{fig:flat-overview} summarizes the trigger-generation pipeline, where the desired class and sampled perturbation remain separate inputs. The target label $t$ specifies the desired misclassification, while the latent code $z$ is intended to condition one perturbation realization for the same source-target pair, with empirical checks used to test whether the generator uses this axis rather than collapsing to one trigger.

The generator state is not a server parameter, so when several malicious clients participate in the same round, they start from the same $\phi^r$, adapt it locally, and average their adapted generator states by local sample count. This sharing requires persistent attacker-side coordination among compromised clients, while the FL server receives only classifier updates in the standard FL format.

\subsection{Latent-Conditioned Trigger Generator}
The generator receives a source image $x$, target label $t$, and latent code $z$. In the conditional-generator variant, the target label is mapped to an embedding $e(t)$ and the latent code is projected to a spatial feature map. The generator then predicts a bounded perturbation,
\begin{align}
    \delta_\phi(x,t,z)
    &=
    \epsilon\,
    \tilde{G}_\phi\left(x,e(t),z\right),\\
    T_\phi(x,t,z)
    &=
    \Pi_{\mathcal{X}}\left(x+\delta_\phi(x,t,z)\right),
\end{align}
where $\tilde{G}_\phi$ has output range $[-1,1]$, $\epsilon$ is the perturbation budget, and $\Pi_{\mathcal{X}}$ clips the image to the valid pixel range. This instantiates the abstract generator from the previous section as $G_\phi(x,t,z)\equiv\delta_\phi(x,t,z)$. For a fixed source image $x$ and target label $t$, the latent code $z$ is the input along which the generator can vary the bounded perturbation without changing the intended class. One generator is reused for every target in the predefined evaluation set $\mathcal{T}$, so FLAT does not claim generalization to labels outside $\mathcal{T}$ unless an all-class setting is explicitly evaluated, and it does not construct a separate trigger or generator for each target.

\subsection{Federated Backdoor Injection}
Before local classifier training, each selected malicious client updates the generator while holding the classifier fixed. Let $q_{\mathcal{T}}(\cdot\mid y)$ denote a target policy supported on $\mathcal{T}\setminus\{y\}$, and let
\[
\mathcal{D}_{k,\mathcal{T}}
=
\{(x,y)\in\mathcal{D}_k:\mathcal{T}\setminus\{y\}\neq\emptyset\}
\]
be the local examples for which at least one active target differs from the source label. For each minibatch from $\mathcal{D}_{k,\mathcal{T}}$, the attacker samples $t\sim q_{\mathcal{T}}(\cdot\mid y)$ and $z\sim p_z$, then optimizes the generator so the frozen classifier predicts $t$ on the triggered image,
\begin{equation}
    \mathcal{L}_{\mathrm{gen},k}^r(\phi \mid \theta^r)
    =
    \mathbb{E}
    \left[
    \mathrm{CE}\left(f_{\theta^r}(T_\phi(x,t,z)),t\right)
    \right].
\end{equation}
where $\mathrm{CE}$ denotes cross-entropy. Adaptation stops once the local backdoor accuracy reaches a preset threshold or the generator-training budget is exhausted, producing $\hat{\phi}_k^r=\mathrm{Adapt}(\phi^r,\mathcal{L}_{\mathrm{gen},k}^r)$. Only $\phi$ changes in this step, and the classifier update sent to the server is produced afterward by the standard local training loop.

After generator adaptation, the malicious client trains the classifier with the same local training loop used by benign clients. For each local sample $(x,y)$, ineligible samples with $\mathcal{T}\setminus\{y\}=\emptyset$ are kept clean, while eligible samples draw $b\sim\mathrm{Bernoulli}(\rho)$. If $b=1$, the attacker samples $t\sim q_{\mathcal{T}}(\cdot\mid y)$ and $z\sim p_z$. The training pair becomes
\begin{equation}
    (\tilde{x},\tilde{y})
    =
    \begin{cases}
    (T_{\hat{\phi}_k^r}(x,t,z),t), & \text{if } b=1,\\
    (x,y), & \text{otherwise}.
    \end{cases}
\end{equation}
The malicious client then trains its local classifier by minimizing
\begin{equation}
    \mathcal{L}_{\mathrm{loc},k}^r(\theta \mid \hat{\phi}_k^r)
    =
    \mathbb{E}_{(\tilde{x},\tilde{y})}
    \left[
    \mathrm{CE}\left(f_\theta(\tilde{x}),\tilde{y}\right)
    \right].
\end{equation}
Here, the expectation over $(\tilde{x},\tilde{y})$ includes the clean-poison mixture process above. In the default setting, FLAT submits $\Delta_k^r=\theta_k^{r,\mathrm{loc}}-\theta^r$ without model replacement, server-code changes, or direct edits to benign-client updates. Any clipping or filtering defense acts on this submitted classifier update in the usual server-side aggregation step, and unless explicitly labeled defense-aware, defense experiments use the same attack protocol while changing only the server rule.

If $\mathcal{M}^r\neq\emptyset$, the attacker updates its generator state with a sample-weighted average,
\begin{equation}
    \phi^{r+1}
    =
    \sum_{k\in\mathcal{M}^r}
    \frac{n_k}{\sum_{j\in\mathcal{M}^r}n_j}\hat{\phi}_k^r,
\end{equation}
where $n_k$ is the local training-set size used for the FedAvg-style attacker-side average. If no malicious client is selected in round $r$, the attacker keeps $\phi^{r+1}=\phi^r$. This averaging occurs only among compromised clients, and no generator state is sent to the FL server. During rounds after the attack window, compromised clients no longer poison local batches or update the generator. Persistence is measured by evaluating the benign-continued global model on inputs generated with the retained attacker-side generator.

\subsection{Target and Latent Sampling}
FLAT treats the target and latent variables separately. The target policy $q_{\mathcal{T}}(\cdot\mid y)$ is supported on $\mathcal{T}\setminus\{y\}$. In the reported experiments, single-target runs use active target set $\{0\}$ and poison only examples with $y\neq0$, while multi-target runs sample from the active set after excluding the source label. Other target schedulers are implementation options and are not compared here. At evaluation time, we condition on a requested target $t$ and draw $z$ according to the evaluation policy.

The latent distribution $p_z$ controls trigger realization. In the main FLAT setting, $z$ is sampled randomly from the chosen latent distribution for each poisoned or evaluated example. Table~\ref{tab:protocol} gives the default latent dimension and generator budget. In the fixed-$z$ control, all examples share one latent code. With $\epsilon$, poison rate, target policy, generator class, and FL schedule held fixed, random-versus-fixed differences in attack activation, worst-target behavior, or trigger diversity are consistent with the role of sampled latent codes in this protocol rather than a different target policy or a separate generator per target.

\section{Experimental Setup}
\label{sec:experimental-setup}
\subsection{Research Questions}
In this paper, we structure the experiments around five questions.
\begin{enumerate}[label=\textbf{Q\arabic*}:, leftmargin=*]
\item \emph{Attack success and clean utility.} Can FLAT induce the chosen target behavior while keeping CA close to a benign FL run trained under the same protocol?
\item \emph{Conditional multi-target behavior.} Does one attacker-side generator support fixed and multi-target goals within a predefined target set without training a separate trigger for each target?
\item \emph{Role of the latent variable.} Under the same perturbation budget and poison rate, how much does random latent sampling change target coverage, per-target ASR, and trigger diversity compared with a fixed-$z$ control?
\item \emph{Behavior under defenses.} Which aggregation or inspection rules reduce ASR, and what CA cost do they introduce?
\item \emph{Persistence and sensitivity.} After malicious clients stop participating, how quickly does ASR decay, and how do results change with the client split, poison rate, target set size, and perturbation budget?
\end{enumerate}

\subsection{Datasets, Models, and FL Protocol}
The primary evaluation is conducted on CIFAR-10, while CIFAR-100 and Tiny-ImageNet are additionally employed to assess whether the proposed protocol generalizes to scenarios with a larger number of classes and an expanded image space~\cite{krizhevsky2009cifar,le2015tinyimagenet}. Table~\ref{tab:datasets} summarizes the data settings. CIFAR training uses random cropping with four-pixel padding, random horizontal flipping, and standard per-channel normalization. Tiny-ImageNet uses the same crop and flip pattern at $64\times64$ resolution with ImageNet normalization. We use independent and identically distributed (IID) for uniform client splits and non-independent and identically distributed (non-IID) for heterogeneous label splits.

\begin{table}[t]
\caption{Datasets and default backbones used in the image-classification experiments.}
\label{tab:datasets}
\centering
\footnotesize
\begin{tabular}{lccc}
\hline
Dataset & Classes & Train/Test & Model \\
\hline
CIFAR-10 & 10 & 50K/10K & ResNet-18-CIFAR \\
CIFAR-100 & 100 & 50K/10K & ResNet-18-CIFAR \\
Tiny-ImageNet & 200 & 100K/10K & ResNet-18-Tiny \\
\hline
\end{tabular}
\end{table}

\begin{table}[!hbt]
\caption{Default FLAT evaluation protocol used unless otherwise specified.}
\label{tab:protocol}
\centering
\footnotesize
\begin{tabular}{p{0.37\linewidth}p{0.53\linewidth}}
\hline
Component & Default setting \\
\hline
Client population & 100 clients, 10 sampled per round \\
Local training & 2 epochs, batch size 64, stochastic gradient descent (SGD) \\
Data split & Dirichlet non-IID, $\alpha=0.5$ \\
Warm start & benign checkpoint after round 199 \\
Attack/post-stop & attack rounds 200--299; benign continuation to round 399 \\
Malicious clients & 10\% of clients, naturally sampled \\
Poisoning & 20\% poisoned data in malicious local batches \\
FLAT generator & conditional autoencoder, $\epsilon=0.03$ \\
Latent policy & random $z$, $d_z=16$; fixed-$z$ ablations \\
Generator update & Adam, learning rate 0.01, up to 200 epochs \\
\hline
\end{tabular}
\end{table}

\begin{table}[t]
\caption{Multi-target operating points used in the CIFAR-10 analysis. Worst-target and coverage values are compared only within the same operating point.}
\label{tab:operating-points}
\centering
\scriptsize
\setlength{\tabcolsep}{3pt}
\renewcommand{\arraystretch}{1.12}
\begin{tabularx}{\columnwidth}{@{}p{0.28\columnwidth}>{\raggedright\arraybackslash}X>{\raggedright\arraybackslash}X@{}}
\hline
Component & Canonical point & Target-scale point \\
\hline
Primary use & Latent ablation, defense heatmap, architecture & Target-set scaling, data heterogeneity \\
Target set & $\mathcal{T}_{\mathrm{can}}=\{0,1,2,3\}$ & Varies with $N_t$ \\
Client split & $\alpha=0.5$ & $\alpha=\{0.3,0.5,0.9\}$ \\
\hline
\multicolumn{3}{@{}l}{\textit{Shared settings}} \\
Attack window & \multicolumn{2}{@{}>{\raggedright\arraybackslash}p{0.69\columnwidth}@{}}{Rounds 200--299} \\
Post-stop window & \multicolumn{2}{@{}>{\raggedright\arraybackslash}p{0.69\columnwidth}@{}}{To round 399} \\
Poison rate & \multicolumn{2}{@{}>{\raggedright\arraybackslash}p{0.69\columnwidth}@{}}{$\rho=0.20$} \\
Perturbation budget & \multicolumn{2}{@{}>{\raggedright\arraybackslash}p{0.69\columnwidth}@{}}{$\epsilon=0.03$} \\
Latent dimension & \multicolumn{2}{@{}>{\raggedright\arraybackslash}p{0.69\columnwidth}@{}}{$d_z=16$} \\
Generator budget & \multicolumn{2}{@{}>{\raggedright\arraybackslash}p{0.69\columnwidth}@{}}{200 gen. epochs, Adam learning rate 0.01} \\
Stopping rule & \multicolumn{2}{@{}>{\raggedright\arraybackslash}p{0.69\columnwidth}@{}}{backdoor accuracy $\ge 0.8$ or budget exhausted} \\
\hline
\end{tabularx}
\end{table}
All image experiments use a ResNet-18 backbone~\cite{he2016resnet} with a $3\times3$ stride-one first convolution and no initial max-pooling layer, which is the CIFAR-style variant used by the codebase. Unless otherwise stated, the FL population has $K=100$ clients. Client data are partitioned with a Dirichlet label distribution using $\alpha=0.5$ as the main non-IID setting. Each seed fixes the client split, sampled clients, and attacker schedule; attack and defense variants reuse the same split when they are compared.

Table~\ref{tab:protocol} gives the default training protocol. Local optimization uses SGD with learning rate 0.1, momentum 0.9, and weight decay $5\times10^{-4}$. The attack begins from the round-199 benign checkpoint, runs for 100 rounds, and is followed by benign continuation to measure post-stop residue. The main fixed-target configuration uses target label 0. Unless otherwise stated, the canonical multi-target setting uses
$\mathcal{T}_{\mathrm{can}}=\{0,1,2,3\}$. For each poisoned or evaluated example, the attacker samples a non-source target from the active target set. The target-scale sweep varies the number of requested targets and is reported as a separate operating point because target difficulty and CA shift across runs.

\textit{Operating points.} Two operating points are used, and worst-target and coverage metrics are compared only within an operating point. The \emph{canonical four-target operating point} ($\mathcal{T}_{\mathrm{can}}=\{0,1,2,3\}$, where $t=2$ is the hardest target) underlies the latent ablation (Table~\ref{tab:additional-flat-ablation}), the architecture check (Table~\ref{tab:model-architectures}), and the defense heatmap (Fig.~\ref{fig:defense-heatmap}). The \emph{target-scale operating point} varies the requested target set under a separate training configuration and underlies Tables~\ref{tab:target-label-size} and~\ref{tab:non-iid-settings}. Because CA and target difficulty shift across these configurations, absolute worst-target and coverage values are not directly comparable across the two. Table~\ref{tab:operating-points} specifies which protocol components are shared and which differ across the two operating points.

\begin{table*}[!hbt]
\caption{Single-target FedAvg attack performance across datasets. CA is reported at the final round, ASR over the attack window, and averages are macro-averages over the three datasets.}
\label{tab:fedavg_attack_summary}
\centering
\scriptsize
\setlength{\tabcolsep}{3.4pt}
\resizebox{\textwidth}{!}{
\begin{tabular}{lcccccccc}
\hline
Method & \multicolumn{2}{c}{CIFAR-10} & \multicolumn{2}{c}{CIFAR-100} & \multicolumn{2}{c}{Tiny-ImageNet} & Avg. CA & Avg. ASR \\
\cline{2-7}
 & CA & ASR & CA & ASR & CA & ASR & & \\
\hline
BadNets & $79.04{\pm}2.84$ & $71.80{\pm}0.28$ & $60.28{\pm}0.45$ & $83.42{\pm}2.23$ & $52.96{\pm}2.39$ & $53.42{\pm}0.65$ & $64.09{\pm}1.89$ & $69.55{\pm}1.05$ \\
DBA & $77.92{\pm}0.87$ & $74.26{\pm}1.31$ & $60.95{\pm}1.20$ & $78.91{\pm}2.43$ & $51.87{\pm}1.46$ & $72.37{\pm}0.15$ & $63.58{\pm}1.18$ & $75.18{\pm}1.30$ \\
Neurotoxin & $80.70{\pm}1.97$ & $79.67{\pm}0.64$ & $61.32{\pm}1.92$ & $88.23{\pm}0.38$ & $51.73{\pm}0.76$ & $82.44{\pm}1.78$ & $64.58{\pm}1.55$ & $83.45{\pm}0.93$ \\
IBA & $81.01{\pm}0.12$ & $90.61{\pm}0.93$ & $59.26{\pm}1.96$ & $95.64{\pm}2.05$ & $52.71{\pm}2.57$ & $91.11{\pm}1.23$ & $64.33{\pm}1.55$ & $92.45{\pm}1.40$ \\
A3FL & $\mathbf{86.53{\pm}0.88}$ & $97.92{\pm}2.54$ & $\mathbf{68.55{\pm}0.30}$ & $99.33{\pm}0.79$ & $\mathbf{58.14{\pm}1.23}$ & $\mathbf{99.44{\pm}2.13}$ & $\mathbf{71.07{\pm}0.80}$ & $\mathbf{98.90{\pm}1.82}$ \\
Venomancer & $78.33{\pm}2.12$ & $\mathbf{99.99{\pm}2.56}$ & $60.20{\pm}2.02$ & $\underline{99.64{\pm}1.52}$ & $\underline{53.85{\pm}1.69}$ & $91.58{\pm}2.13$ & $64.13{\pm}1.94$ & $97.07{\pm}2.07$ \\
FLAT & $\underline{85.01{\pm}1.26}$ & $\underline{99.49{\pm}1.48}$ & $\underline{63.06{\pm}0.83}$ & $\mathbf{99.66{\pm}1.64}$ & $53.82{\pm}1.75$ & $\underline{94.10{\pm}0.70}$ & $\underline{67.30{\pm}1.28}$ & $\underline{97.75{\pm}1.27}$ \\
\hline
\end{tabular}
}
\end{table*}

\subsection{Baselines, Defenses, and Metrics}

The attack baselines include fixed-trigger attacks and FL-specific backdoors such as BadNets~\cite{gu2017badnets}, DBA~\cite{xie2020dba}, Neurotoxin~\cite{zhang2022neurotoxin}, IBA~\cite{nguyen2023iba}, and A3FL~\cite{zhang2023a3fl}. Venomancer~\cite{nguyen2024venomancer} is the closest target-on-demand reference; we report its numbers only when the run uses the same client split, model, and FL schedule.

Defense experiments begin with FedAvg and then change one server rule at a time. The robust aggregation group includes coordinate median~\cite{yin2018byzantine}, trimmed mean~\cite{yin2018byzantine}, Krum~\cite{blanchard2017krum}, Multi-Krum~\cite{blanchard2017krum}, and RFA~\cite{pillutla2022rfa}. Similarity-, trust-, clustering-, and inspection-based defenses include FoolsGold~\cite{fung2020foolsgold}, FLTrust~\cite{cao2021fltrust}, FLAME~\cite{nguyen2022flame}, AlignIns~\cite{xu2025alignins}, MARS~\cite{wan2025mars}, BackdoorIndicator~\cite{li2024backdoorindicator}, FLIP~\cite{zhang2023flip}, and FLShield~\cite{kabir2024flshield}. Keeping the client training code and attack schedule fixed helps attribute differences mainly to the server rule; when a baseline requires a different schedule, we treat that case separately.

The main tables report CA and ASR. Where the run is multi-target and the table has space, we also report Avg ASR, worst-target ASR, AllTargets@80, and post-stop ASR, so a strong average cannot hide a failed target or a rapid post-attack decay.

\section{Results and Analysis}
\label{sec:results}

\textbf{Evaluation lens.}
The central question is not only whether FLAT achieves high ASR, but whether standard HFL auditing would miss a broader hidden behavior surface. We therefore evaluate FLAT through clean utility plus six reliability measurements: (i) target-wise behavior, (ii) worst-target behavior, (iii) target coverage, (iv) latent-trigger variation, (v) defense response, and (vi) post-stop persistence. CA measures clean utility. Mean ASR measures average attack activation. Per-target ASR shows which target modes are active. Worst-target ASR and AllTargets@80 measure whether the requested targets are jointly covered rather than only successful on average. Defense-response results show whether server-side rules remove the behavior uniformly or only reshape the target-conditioned surface. Post-stop ASR measures whether the hidden behavior remains after malicious clients stop participating. AllTargets@80 is the fraction of requested targets whose ASR is at least $80\%$.

The subsections answer Q1--Q5 in order, moving from single-target utility to target-set scaling, latent variation, defense response, and post-stop or protocol sensitivity.

\begin{table*}[!hbt]
\caption{CIFAR-10 FLAT sensitivity to target-label set size under the target-scale operating point. The target subset is reported explicitly because target difficulty is non-uniform. AllTargets@80 is the fraction of target labels whose ASR is at least $80\%$.}
\label{tab:target-label-size}
\centering
\scriptsize
\setlength{\tabcolsep}{3.5pt}
\begin{tabular}{llccccc}
\hline
Target Set Size & Target Labels & CA & Avg ASR & Worst ASR & AllTargets@80 & Post-stop ASR \\
\hline
$N_t=1$  & $\{0\}$ & $\mathbf{85.01{\pm}1.26}$ & $\mathbf{99.49{\pm}1.48}$ & $\mathbf{99.49{\pm}1.48}$ & $\mathbf{100.00}$ & $32.34{\pm}1.54$ \\
$N_t=2$  & $\{0,1\}$ & $76.71{\pm}1.21$ & $\underline{98.11{\pm}1.67}$ & $\underline{97.01{\pm}0.88}$ & $\mathbf{100.00}$ & $24.40{\pm}1.25$ \\
$N_t=4$  & $\{0,1,2,3\}$ & $\underline{80.39{\pm}1.12}$ & $91.80{\pm}1.35$ & $88.82{\pm}0.97$ & $\mathbf{100.00}$ & $\mathbf{51.80{\pm}1.68}$ \\
$N_t=5$  & $\{0,1,2,3,4\}$ & $77.15{\pm}1.44$ & $95.12{\pm}1.26$ & $92.43{\pm}0.35$ & $\mathbf{100.00}$ & $\underline{45.01{\pm}1.90}$ \\
$N_t=8$  & $\{0,1,2,3,4,5,6,7\}$ & $76.82{\pm}0.74$ & $93.23{\pm}0.49$ & $77.81{\pm}1.81$ & $87.50$ & $36.74{\pm}1.12$ \\
$N_t=10$ & $\{0,1,2,3,4,5,6,7,8,9\}$ & $77.24{\pm}1.67$ & $89.79{\pm}1.39$ & $56.91{\pm}0.80$ & $\underline{90.00}$ & $46.60{\pm}0.52$ \\
\hline
\end{tabular}
\end{table*}

\subsection{Single-Target Setting}

\textbf{FedAvg performance.}
Table~\ref{tab:fedavg_attack_summary} summarizes the no-defense single-target setting across the three datasets. The FLAT row reaches $99.49\%$ attack-window ASR on CIFAR-10, $99.66\%$ on CIFAR-100, and $94.10\%$ on Tiny-ImageNet, with final CA values of $85.01\%$, $63.06\%$, and $53.82\%$. The same table bounds the claim because A3FL has the highest macro CA and macro ASR, and Venomancer nearly saturates CIFAR-10. FLAT's relevant distinction is therefore conditional generation, where one attacker-side generator $G(x,t,z)$ uses the target label and latent code to produce latent-selected perturbations while staying competitive in the FedAvg reference setting.

\subsection{Target-Space Scalability}

\begin{figure}[!hbt]
\centering
\includegraphics[width=0.92\linewidth]{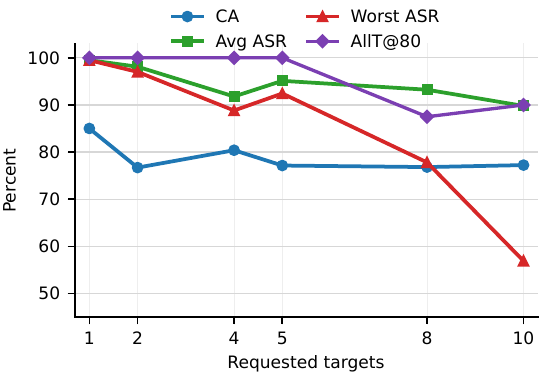}
\caption{CIFAR-10 target-set scaling for FLAT, showing CA, average ASR, worst-target ASR, and AllTargets@80 as the requested target set grows.}
\label{fig:target-scale}
\end{figure}

\textbf{Scaling target set size.}
Table~\ref{tab:target-label-size} and Fig.~\ref{fig:target-scale} evaluate whether one conditional FLAT generator can support more requested target labels within the target-scale setting. In these selected CIFAR-10 rows, $N_t=1,2,4,5$ all reach AllTargets@80 $=100\%$. The larger rows expose the limit. Avg ASR remains $93.23\%$ for $N_t=8$ and $89.79\%$ for $N_t=10$, but worst-target ASR drops to $77.81\%$ and $56.91\%$. The plot makes the same separation visible, with broad activation remaining high while weakest-target reliability is where the target-set failure appears.

\subsection{Latent Trigger Variation and Component Ablations}

\begin{table*}[t]
\caption{CIFAR-10 FLAT component ablations under the canonical four-target operating point.}
\label{tab:additional-flat-ablation}
\centering
\scriptsize
\setlength{\tabcolsep}{3.5pt}
\begin{tabular}{llccccc}
\hline
Variant & Change & CA & Avg ASR & Worst ASR & AllTargets@80 & Post-stop ASR \\
\hline
Fixed $z$    & shared $z$    & 70.20$\pm$1.12 & 81.20$\pm$1.87 & 40.68$\pm$2.61 & 75.00 & 32.31$\pm$0.93 \\
Random $z$   & sampled $z$   & 80.39$\pm$1.12 & 91.80$\pm$1.35 & 88.82$\pm$0.97 & 100.00 & 51.80$\pm$1.68 \\
High $\epsilon$ & $\epsilon=0.05$ & 69.88$\pm$1.26 & 84.23$\pm$1.73 & 43.20$\pm$2.35 & 75.00 & 53.05$\pm$2.81 \\
Low $\rho$   & $\rho=0.05$   & 72.32$\pm$0.91 & 77.06$\pm$2.23 & 40.02$\pm$1.14 & 75.00 & 34.65$\pm$1.52 \\
Short train  & 50 gen. epochs  & 69.39$\pm$1.37 & 85.67$\pm$1.94 & 63.90$\pm$2.17 & 75.00 & 26.92$\pm$0.76 \\
\hline
\end{tabular}
\end{table*}

\textbf{Latent ablation.}
Table~\ref{tab:additional-flat-ablation} uses the canonical setting introduced in Table~\ref{tab:operating-points} and separates the latent variable from the target-conditioning mechanism. The fixed latent-code control reuses one code for all poisoned examples, while the random latent-code row samples a fresh code for the same $(x,t)$ pair. Avg ASR moves from $81.20\%$ to $91.80\%$, and worst-target ASR rises from $40.68\%$ to $88.82\%$, while AllTargets@80 increases from $75\%$ to $100\%$. The latent input therefore changes weakest-target coverage in this setting.

\textbf{Training and budget controls.}
The remaining rows in Table~\ref{tab:additional-flat-ablation} move utility, activation, and durability in different directions. Raising the perturbation budget to $\epsilon=0.05$ increases post-stop ASR to $53.05\%$ while keeping AllTargets@80 at $75\%$. Lowering the poison rate to $\rho=0.05$ improves CA to $72.32\%$ while reducing average ASR to $77.06\%$, so the lower-poison setting gives up attack strength. The short-train row separates immediate activation from durability. Among the three training and budget controls, it has the strongest worst-target ASR, $63.90\%$, and the lowest post-stop ASR, $26.92\%$, which means the attack-window peak alone would overstate persistence.

\begin{figure}[t]
\centering
\includegraphics[width=0.92\linewidth]{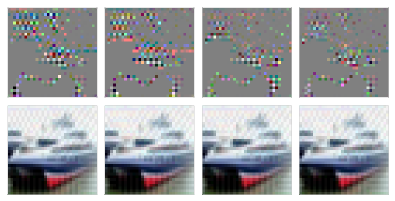}
\caption{Latent-conditioned trigger realizations for the same source image and target label.}
\label{fig:latent-grid}
\end{figure}

\begin{figure}[t]
\centering
\includegraphics[width=0.92\linewidth]{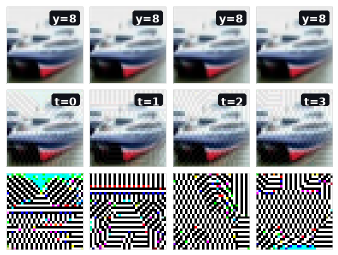}
\caption{Clean images, triggered images, and amplified perturbations across target labels and latent samples.}
\label{fig:trigger-grid}
\end{figure}

\textbf{Visual latent evidence.}
Figures~\ref{fig:latent-grid} and~\ref{fig:trigger-grid} provide the visual counterpart to Table~\ref{tab:additional-flat-ablation}. They are sanity checks, not perceptual stealth evidence. The grids show that FLAT samples a family of target-conditioned perturbations rather than a single fixed patch.

\begin{table*}[!hbt]
\caption{CIFAR-10 defense response matrix. Each attack reports CA and ASR.}
\label{tab:cifar10_full_matrix}
\centering
\tiny
\setlength{\tabcolsep}{2.0pt}
\resizebox{\textwidth}{!}{%
\begin{tabular}{l*{6}{cc}}
\hline
Defense & \multicolumn{2}{c}{BadNets} & \multicolumn{2}{c}{Neurotoxin} & \multicolumn{2}{c}{IBA} & \multicolumn{2}{c}{A3FL} & \multicolumn{2}{c}{Venomancer} & \multicolumn{2}{c}{FLAT} \\
\cline{2-13}
 & CA & ASR & CA & ASR & CA & ASR & CA & ASR & CA & ASR & CA & ASR \\
\hline
FedAvg & 79.04$\pm$2.8 & 71.80$\pm$0.3 & 80.70$\pm$2.0 & 79.67$\pm$0.6 & 81.01$\pm$0.1 & 90.61$\pm$0.9 & 86.53$\pm$0.9 & 97.92$\pm$2.5 & 78.33$\pm$2.1 & 99.99$\pm$2.6 & 85.01$\pm$1.3 & 99.49$\pm$1.5 \\
Median & 89.35$\pm$0.4 & 96.03$\pm$1.2 & 89.44$\pm$1.6 & 94.68$\pm$1.4 & 89.58$\pm$0.7 & 99.57$\pm$0.3 & 86.58$\pm$1.9 & 49.53$\pm$0.7 & 79.97$\pm$1.7 & 51.00$\pm$1.2 & 89.46$\pm$1.6 & 96.64$\pm$1.5 \\
TrimMean & 89.75$\pm$1.1 & 96.16$\pm$1.9 & $\mathbf{89.75{\pm}1.7}$ & 95.94$\pm$1.2 & $\underline{89.73{\pm}1.7}$ & 99.57$\pm$1.1 & 86.74$\pm$1.4 & 50.98$\pm$0.1 & $\underline{80.55{\pm}0.2}$ & 38.93$\pm$0.5 & $\underline{89.64{\pm}0.2}$ & 96.95$\pm$0.6 \\
Krum & 83.07$\pm$1.3 & 3.56$\pm$0.7 & 83.15$\pm$0.5 & 1.47$\pm$1.9 & 82.98$\pm$1.3 & 95.06$\pm$1.2 & 82.50$\pm$0.3 & $\mathbf{6.49{\pm}1.5}$ & 50.17$\pm$2.0 & 12.89$\pm$1.3 & 82.85$\pm$1.1 & 88.03$\pm$1.4 \\
Multi-Krum & $\underline{89.76{\pm}1.7}$ & 96.06$\pm$1.6 & 89.69$\pm$0.6 & 97.11$\pm$0.5 & 89.69$\pm$0.4 & 99.64$\pm$1.9 & $\mathbf{87.06{\pm}1.8}$ & 62.24$\pm$0.6 & 73.00$\pm$1.8 & 12.11$\pm$0.9 & 89.52$\pm$0.5 & 97.39$\pm$0.5 \\
FoolsGold & 89.34$\pm$1.1 & 89.14$\pm$0.5 & 89.64$\pm$0.8 & 97.41$\pm$0.4 & 89.52$\pm$2.0 & 99.60$\pm$1.0 & 79.59$\pm$0.2 & 62.94$\pm$0.1 & 77.85$\pm$1.6 & 31.21$\pm$0.8 & 89.52$\pm$0.1 & 96.85$\pm$0.8 \\
RFA & $\mathbf{89.95{\pm}2.0}$ & 96.22$\pm$1.1 & $\underline{89.70{\pm}0.0}$ & 97.04$\pm$1.4 & $\mathbf{89.93{\pm}1.4}$ & 99.82$\pm$1.1 & $\underline{86.98{\pm}0.5}$ & 66.68$\pm$1.3 & 78.90$\pm$0.9 & 27.44$\pm$1.9 & $\mathbf{89.82{\pm}1.8}$ & 96.60$\pm$0.5 \\
FLAME & 87.12$\pm$1.0 & 80.23$\pm$0.4 & 87.19$\pm$0.6 & 95.02$\pm$1.3 & 87.30$\pm$1.2 & 99.16$\pm$0.3 & 86.64$\pm$1.5 & 56.93$\pm$1.1 & 65.79$\pm$0.0 & 60.53$\pm$0.7 & 87.42$\pm$0.0 & 68.47$\pm$1.9 \\
FLTrust & 85.78$\pm$1.8 & 2.24$\pm$1.7 & 85.91$\pm$1.8 & 2.59$\pm$1.9 & 85.86$\pm$0.2 & 91.52$\pm$1.0 & 85.70$\pm$0.1 & $\underline{13.04{\pm}1.5}$ & $\mathbf{83.16{\pm}0.9}$ & 79.50$\pm$1.1 & 86.01$\pm$0.5 & 88.94$\pm$1.7 \\
AlignIns & 71.82$\pm$0.8 & $\mathbf{0.71{\pm}0.4}$ & 57.80$\pm$0.4 & $\underline{0.59{\pm}0.6}$ & 67.73$\pm$2.0 & $\mathbf{1.11{\pm}1.3}$ & 86.04$\pm$0.9 & 93.23$\pm$1.0 & 66.95$\pm$0.7 & 54.42$\pm$1.2 & 64.13$\pm$0.5 & $\mathbf{18.00{\pm}0.4}$ \\
MARS & 74.29$\pm$0.1 & $\underline{1.52{\pm}1.3}$ & 73.53$\pm$1.7 & 2.23$\pm$0.1 & 78.45$\pm$0.5 & 26.98$\pm$1.3 & 86.71$\pm$0.4 & 65.69$\pm$0.3 & 75.57$\pm$0.9 & 20.56$\pm$1.6 & 77.06$\pm$1.6 & 32.40$\pm$0.4 \\
BD-Indic. & 79.73$\pm$0.2 & 2.46$\pm$0.9 & 77.99$\pm$1.5 & 2.87$\pm$1.4 & 77.98$\pm$2.0 & $\underline{7.87{\pm}0.2}$ & 86.67$\pm$0.8 & 68.31$\pm$0.7 & 75.55$\pm$0.4 & 20.86$\pm$0.9 & 79.61$\pm$0.8 & $\underline{19.03{\pm}0.6}$ \\
FLIP & 75.27$\pm$0.5 & 2.14$\pm$1.9 & 79.02$\pm$1.1 & 1.66$\pm$0.1 & 76.20$\pm$2.0 & 8.86$\pm$1.7 & 86.82$\pm$1.9 & 60.32$\pm$1.9 & 78.33$\pm$1.0 & $\underline{7.52{\pm}0.4}$ & 79.62$\pm$0.8 & 30.61$\pm$0.1 \\
FLShield & 72.80$\pm$0.8 & 6.47$\pm$2.0 & 72.60$\pm$0.9 & $\mathbf{0.40{\pm}0.8}$ & 70.86$\pm$1.9 & 9.88$\pm$2.0 & 85.68$\pm$1.1 & 62.82$\pm$1.4 & 71.14$\pm$1.9 & $\mathbf{7.12{\pm}1.2}$ & 80.12$\pm$1.1 & 41.78$\pm$1.5 \\
\hline
\end{tabular}
}
\end{table*}

\begin{figure*}[t]
\centering
\includegraphics[width=0.94\textwidth]{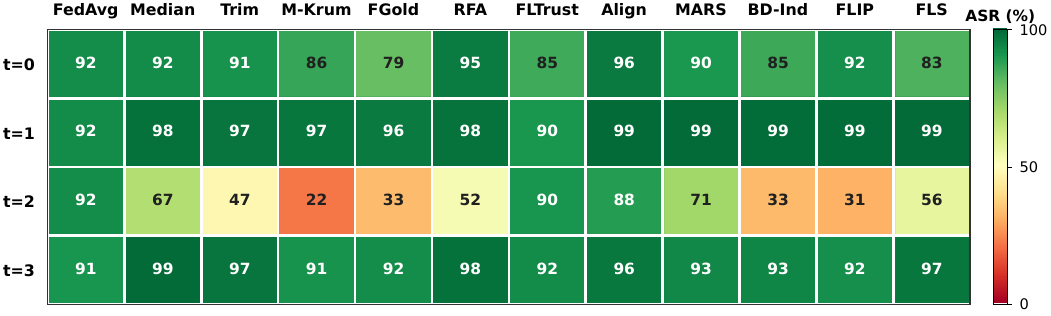}
\caption{Per-target ASR under server-side defenses for CIFAR-10 multi-target FLAT. Each row corresponds to a defense and each column to a target label, with color indicating target-wise ASR. The heatmap reveals non-uniform defense responses, where some target modes are suppressed while others remain highly active.}
\label{fig:defense-heatmap}
\end{figure*}

\subsection{Single-Target Defense Response}

Table~\ref{tab:cifar10_full_matrix} reports output-level defense response for the single-target CIFAR-10 setting. The FLAT column stays above $96\%$ ASR under Median, TrimMean, Multi-Krum, FoolsGold, and RFA. Krum and FLTrust reduce it, but the remaining ASR is still $88.03\%$ and $88.94\%$. The inspection-oriented rows are sharper. AlignIns, BD-Indic., FLIP, MARS, and FLShield reduce FLAT to $18.00\%$, $19.03\%$, $30.61\%$, $32.40\%$, and $41.78\%$ ASR, respectively. These defenses inspect direction, clustering, sparsity, or validation behavior, and their CA costs differ. The single-target defense matrix is not update-space evidence. It shows the narrower point that robust aggregation can still leave high single-target ASR in the final global model. In the separate multi-target setting, Fig.~\ref{fig:defense-heatmap} adds a target-wise qualification, since a defense can reduce the aggregate number while leaving one target mode alive.

\subsection{Multi-Target Defense Response}

\textbf{Per-target defense surface.}
Figure~\ref{fig:defense-heatmap} uses the canonical CIFAR-10 four-target setting from Table~\ref{tab:operating-points}. Under FedAvg, all four target modes remain active, with target-wise mean ASR between $90.62\%$ and $92.40\%$. The same operating point has $91.80\%$ average ASR and $88.82\%$ seed-level worst-target ASR in the $N_t=4$ row of Table~\ref{tab:target-label-size}. The defended rows show a more uneven response. Median, TrimMean, Multi-Krum, FoolsGold, RFA, BD-Indic., FLIP, and FLShield hit $t=2$ hardest, while $t=0$, $t=1$, and $t=3$ often stay high. Read next to Table~\ref{tab:cifar10_full_matrix}, this gap is important. In the single-target run, inspection-oriented rules push FLAT down to $18.00$--$41.78\%$ ASR. In the multi-target run, several of the same rule families still leave multiple targets above $80\%$. A single ASR number misses that failure mode. Some targets are suppressed, others remain exposed, and single-target defense success can look stronger than the protection it gives against a target-conditioned attack.

\begin{table}[t]
\caption{CIFAR-10 FLAT results under representative client data heterogeneity settings. Dirichlet $\alpha$ controls non-IID severity, with smaller values giving more skewed client label distributions.}
\label{tab:non-iid-settings}
\centering
\scriptsize
\setlength{\tabcolsep}{2.2pt}
\resizebox{\columnwidth}{!}{%
\begin{tabular}{lcccc}
\hline
Setting & CA & ASR & Worst ASR & Post-stop ASR \\
\hline
IID & 90.44$\pm$0.72 & 99.75$\pm$0.44 & 99.56$\pm$0.58 & 70.22$\pm$1.14 \\
$\alpha=0.3$ & 77.74$\pm$0.95 & 95.09$\pm$0.66 & 88.49$\pm$1.51 & 37.29$\pm$1.73 \\
$\alpha=0.5$ & 80.39$\pm$1.12 & 91.80$\pm$1.35 & 88.82$\pm$0.97 & 51.80$\pm$1.68 \\
$\alpha=0.9$ & 87.65$\pm$0.79 & 93.16$\pm$0.91 & 88.62$\pm$1.26 & 45.94$\pm$1.09 \\
\hline
\end{tabular}%
}
\end{table}

\subsection{Data Heterogeneity and Architecture Sensitivity}

\textbf{Data heterogeneity.}
Table~\ref{tab:non-iid-settings} asks whether FLAT's behavior is an artifact of a convenient CIFAR-10 client split. The answer is largely no for attack activation. ASR stays above $91\%$ for every Dirichlet partition and reaches $99.75\%$ under IID, while Worst ASR remains between $88.49\%$ and $88.82\%$ for the Dirichlet rows. The partition matters more for the quantities around the attack. Post-stop ASR is $37.29\%$ at $\alpha=0.3$, $51.80\%$ at $\alpha=0.5$, and $45.94\%$ at $\alpha=0.9$, while IID rises to $70.22\%$. Heterogeneity therefore changes how clean utility and persistence travel with the backdoor, even when the attack itself remains active.

\textbf{Architecture sensitivity.}
Table~\ref{tab:model-architectures} reports architecture runs under the canonical four-target operating point. The ResNet-18 row is the canonical FedAvg reference and matches the $N_t=4$ row in Table~\ref{tab:target-label-size}. ResNet-34 and VGG-19 lower both ASR and Worst ASR under the same protocol. This is a sensitivity check, not an architecture-agnostic claim. Changing the backbone changes CA and weakest-target behavior even when the attack schedule is fixed.

\begin{table}[!hbt]
\caption{Selected CIFAR-10 FLAT architecture sensitivity results under the canonical four-target operating point. All rows use the same FL split, target-label set, attacker budget, and attack window.}
\label{tab:model-architectures}
\centering
\scriptsize
\setlength{\tabcolsep}{3.2pt}
\resizebox{\columnwidth}{!}{%
\begin{tabular}{lcccc}
\hline
Architecture & Params & CA & ASR & Worst ASR \\
\hline
ResNet-18 & 11.17M & 80.39$\pm$1.12 & 91.80$\pm$1.35 & 88.82$\pm$0.97 \\
ResNet-34 & 21.28M & 64.98$\pm$1.24 & 68.97$\pm$1.83 & 38.80$\pm$0.77 \\
VGG-19 & 20.29M & 72.97$\pm$1.51 & 71.04$\pm$1.42 & 56.62$\pm$0.58 \\
\hline
\end{tabular}
}
\end{table}

\begin{figure}[!hbt]
\centering
\includegraphics[width=0.94\linewidth]{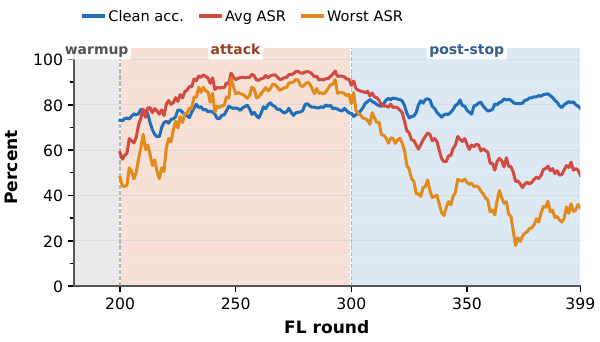}
\caption{Durability over FL rounds for CIFAR-10 multi-target FLAT, showing benign warmup, attack window, and post-stop benign continuation.}
\label{fig:persistence}
\end{figure}

\textbf{Post-stop persistence.}
Figure~\ref{fig:persistence} tracks CIFAR-10 multi-target FLAT through benign warmup, the attack window, and the post-stop continuation. By the end of the attack window, Avg ASR and worst-target ASR have both risen, but after malicious clients leave, the two curves decay at different rates and worst-target ASR falls faster. The run therefore leaves residual behavior after the attack stops, while continued benign training erodes the least reliable target first.

\section{Discussion, Scope, and Limitations}
\label{sec:discussion}
\subsection{Implications for HFL Reliability Audits}
FLAT shifts reliability evaluation from a single trigger-target association toward a target-conditioned family of possible failures. Fixed-trigger backdoors are often summarized by one trigger pattern and one target label, but the FLAT setting evaluates the same requested target under multiple sampled realizations. The fixed- and random-latent controls in Table~\ref{tab:additional-flat-ablation}, together with the visual grids in Figs.~\ref{fig:latent-grid} and~\ref{fig:trigger-grid}, support this audit concern while stopping short of a complete quantitative characterization of trigger-mode diversity. An audit based only on clean validation or one sampled trigger may understate the hidden behavior retained by the trained model.

Aggregate metrics create a related blind spot. Table~\ref{tab:target-label-size} and Fig.~\ref{fig:target-scale} show why average ASR and worst-target ASR should be read together, since the target set can keep a high average while one target remains substantially weaker, and Fig.~\ref{fig:persistence} adds the temporal question of what happens after malicious participation stops. In this paper, target-wise ASR, worst-target ASR, target coverage, latent-sampled evaluation, defense response, and post-stop ASR are used as security-reliability stress-test indicators for the evaluated HFL setting rather than estimates of deployment failure probability. Within the predefined target set and threat model, these indicators expose weak-target and post-stop behavior that a single aggregate ASR loses.

Defense evaluation inherits the same measurement issue. Server-side rules may inspect update norms, coordinate outliers, client similarity, clustering structure, or trusted-data alignment, and those signals can reduce some backdoor behaviors, yet output-level testing is still needed because such signals alone do not empirically rule out target-conditioned behavior in the final model. The single-target defense matrix in Table~\ref{tab:cifar10_full_matrix} and the four-target heatmap in Fig.~\ref{fig:defense-heatmap} show this issue on CIFAR-10, where defenses can suppress some target modes more strongly than others. Rather than showing that existing defenses categorically fail against FLAT, these results support the narrower recommendation that defense evaluation include behavioral tests across requested targets and latent samples, while update-space diagnostics remain future work.

\subsection{Scope and Limitations}
The experiments focus on HFL image classification with CIFAR-10, CIFAR-100, and Tiny-ImageNet. The cross-dataset evidence in Table~\ref{tab:fedavg_attack_summary} primarily establishes single-target FedAvg viability, whereas the richer analyses of target scaling, latent ablations, defense heatmaps, non-IID partitions, architecture sensitivity, and persistence are centered on CIFAR-10. These controlled benchmarks help isolate target-conditioned latent-trigger behavior, but they do not establish transfer to text models, vertical FL, graph FL, personalized FL, physical triggers, or production deployments. Secure aggregation, client churn, dropout, compression, differential privacy noise, limited audit data, device heterogeneity, and large-scale partial participation may each change what a client observes, what an attacker can edit, and what the server can validate.

Threat-model comparability is also limited by the attacker assumptions. FLAT assumes that the attacker controls a subset of clients, can alter their local training procedure, and can maintain attacker-side generator state across rounds, which implies persistent malicious identities or out-of-band coordination among compromised clients. The server is not compromised and does not receive the generator. Baselines that force-include malicious clients or alter the participation rule should be reported separately from the sampled-client FLAT setting, and defense-aware variants should be marked separately from fixed-protocol evaluations because adaptive defense-specific optimization is outside scope unless explicitly stated.

The attack introduces attacker-side computational overhead because compromised clients optimize a generator alongside the classifier update. The magnitude of this overhead depends on the generator architecture, latent dimension, number of target labels, perturbation budget, and stopping rule in Table~\ref{tab:protocol}, although this paper does not report a separate runtime or energy benchmark. The visual examples in Figs.~\ref{fig:latent-grid} and~\ref{fig:trigger-grid} provide only a preliminary perceptibility check, not a substitute for task-specific perceptual metrics, a human study, usability evaluation, or physical-world validation.

FLAT is a dual-use method and is presented here as an authorized reliability stress test for FL systems, not as deployment guidance for compromising real systems. Implementations, artifacts, and benchmark use should be limited to settings where the evaluator has permission to test, and reports should include enough context to make the evaluation auditable, including the client split, attack window, malicious-client sampling rule, target policy, latent sampling rule, perturbation budget, and defense diagnostics. Those reporting details support reproducibility while keeping the intended use within controlled research, red-team, and defense-evaluation settings.
\section{Conclusion}
\label{sec:conclusion}
FLAT turns a familiar HFL backdoor question into a harder audit problem because the target label is no longer tied to one visible trigger. Malicious clients keep a generator, $G(x,t,z)$, that produces a family of perturbations from the input, target, and latent code, so evaluation has to follow behavior across targets and samples. In the image-classification experiments, FLAT remains strong under FedAvg on different datasets, while the multi-target runs show what mean ASR misses. Latent sampling changes the weakest-target coverage, larger target sets expose worse target degradation, and post-stop benign training erodes the weakest target before removing all residual behavior. The defense and architecture studies keep the claim bounded. Some inspection rules sharply reduce single-target ASR, server defenses suppress target modes unevenly, and different backbones weaken both ASR and Worst ASR. A useful FL backdoor report therefore needs target-wise ASR, Worst ASR, target coverage, post-stop persistence, and defense response alongside CA, instead of a single mean ASR and one known trigger.

\bibliographystyle{IEEEtran}
\bibliography{references}

\begin{thebibliography}{10}
\providecommand{\url}[1]{#1}
\csname url@samestyle\endcsname
\providecommand{\newblock}{\relax}
\providecommand{\bibinfo}[2]{#2}
\providecommand{\BIBentrySTDinterwordspacing}{\spaceskip=0pt\relax}
\providecommand{\BIBentryALTinterwordstretchfactor}{4}
\providecommand{\BIBentryALTinterwordspacing}{\spaceskip=\fontdimen2\font plus
\BIBentryALTinterwordstretchfactor\fontdimen3\font minus
  \fontdimen4\font\relax}
\providecommand{\BIBforeignlanguage}[2]{{%
\expandafter\ifx\csname l@#1\endcsname\relax
\typeout{** WARNING: IEEEtran.bst: No hyphenation pattern has been}%
\typeout{** loaded for the language `#1'. Using the pattern for}%
\typeout{** the default language instead.}%
\else
\language=\csname l@#1\endcsname
\fi
#2}}
\providecommand{\BIBdecl}{\relax}
\BIBdecl

\bibitem{bagdasaryan2020backdoorfl}
\BIBentryALTinterwordspacing
E.~Bagdasaryan, A.~Veit, Y.~Hua, D.~Estrin, and V.~Shmatikov, ``How to backdoor
  federated learning,'' in \emph{Proceedings of the International Conference on
  Artificial Intelligence and Statistics (AISTATS)}, ser. Proceedings of
  Machine Learning Research, vol. 108.\hskip 1em plus 0.5em minus 0.4em\relax
  PMLR, 2020, pp. 2938--2948. [Online]. Available:
  \url{https://proceedings.mlr.press/v108/bagdasaryan20a.html}
\BIBentrySTDinterwordspacing

\bibitem{xie2020dba}
\BIBentryALTinterwordspacing
C.~Xie, K.~Huang, P.-Y. Chen, and B.~Li, ``{DBA}: Distributed backdoor attacks
  against federated learning,'' in \emph{Proceedings of the International
  Conference on Learning Representations (ICLR)}, 2020. [Online]. Available:
  \url{https://openreview.net/forum?id=rkgyS0VFvr}
\BIBentrySTDinterwordspacing

\bibitem{zhang2022neurotoxin}
\BIBentryALTinterwordspacing
Z.~Zhang, A.~Panda, L.~Song, Y.~Yang, M.~W. Mahoney, P.~Mittal, K.~Ramchandran,
  and J.~Gonzalez, ``Neurotoxin: Durable backdoors in federated learning,'' in
  \emph{Proceedings of the International Conference on Machine Learning
  (ICML)}, ser. Proceedings of Machine Learning Research, vol. 162.\hskip 1em
  plus 0.5em minus 0.4em\relax PMLR, 2022, pp. 26\,429--26\,446. [Online].
  Available: \url{https://proceedings.mlr.press/v162/zhang22w.html}
\BIBentrySTDinterwordspacing

\bibitem{nguyen2023iba}
\BIBentryALTinterwordspacing
D.~T. Nguyen, T.~Nguyen, T.~A. Tran, K.~D. Doan, and K.-S. Wong, ``{IBA}:
  Towards irreversible backdoor attacks in federated learning,'' in
  \emph{Advances in Neural Information Processing Systems (NeurIPS)}, 2023.
  [Online]. Available:
  \url{https://proceedings.neurips.cc/paper_files/paper/2023/hash/d0c6bc641a56bebee9d985b937307367-Abstract-Conference.html}
\BIBentrySTDinterwordspacing

\bibitem{zhang2023a3fl}
\BIBentryALTinterwordspacing
H.~Zhang, J.~Jia, J.~Chen, L.~Lin, and D.~Wu, ``{A3FL}: Adversarially adaptive
  backdoor attacks to federated learning,'' in \emph{Advances in Neural
  Information Processing Systems (NeurIPS)}, 2023. [Online]. Available:
  \url{https://openreview.net/forum?id=S6ajVZy6FA}
\BIBentrySTDinterwordspacing

\bibitem{doan2022marksman}
\BIBentryALTinterwordspacing
K.~D. Doan, Y.~Lao, and P.~Li, ``Marksman backdoor: Backdoor attacks with
  arbitrary target class,'' in \emph{Advances in Neural Information Processing
  Systems (NeurIPS)}, 2022. [Online]. Available:
  \url{https://proceedings.neurips.cc/paper_files/paper/2022/hash/fa0126bb7ebad258bf4ffdbbac2dd787-Abstract-Conference.html}
\BIBentrySTDinterwordspacing

\bibitem{nguyen2024venomancer}
\BIBentryALTinterwordspacing
S.~Nguyen, T.~Nguyen, K.~D. Doan, and K.-S. Wong, ``Venomancer: Towards
  imperceptible and target-on-demand backdoor attacks in federated learning,''
  2024. [Online]. Available: \url{https://arxiv.org/abs/2407.03144}
\BIBentrySTDinterwordspacing

\bibitem{mcmahan2017fedavg}
\BIBentryALTinterwordspacing
B.~McMahan, E.~Moore, D.~Ramage, S.~Hampson, and B.~A. y~Arcas,
  ``Communication-efficient learning of deep networks from decentralized
  data,'' in \emph{Proceedings of the International Conference on Artificial
  Intelligence and Statistics (AISTATS)}, ser. Proceedings of Machine Learning
  Research, vol.~54.\hskip 1em plus 0.5em minus 0.4em\relax PMLR, 2017, pp.
  1273--1282. [Online]. Available:
  \url{https://proceedings.mlr.press/v54/mcmahan17a.html}
\BIBentrySTDinterwordspacing

\bibitem{li2020fedprox}
\BIBentryALTinterwordspacing
T.~Li, A.~K. Sahu, M.~Zaheer, M.~Sanjabi, A.~Talwalkar, and V.~Smith,
  ``Federated optimization in heterogeneous networks,'' in \emph{Proceedings of
  Machine Learning and Systems (MLSys)}, vol.~2, 2020, pp. 429--450. [Online].
  Available:
  \url{https://proceedings.mlsys.org/paper/2020/hash/1f5fe83998a09396ebe6477d9475ba0c-Abstract.html}
\BIBentrySTDinterwordspacing

\bibitem{karimireddy2020scaffold}
\BIBentryALTinterwordspacing
S.~P. Karimireddy, S.~Kale, M.~Mohri, S.~Reddi, S.~Stich, and A.~T. Suresh,
  ``{SCAFFOLD}: Stochastic controlled averaging for federated learning,'' in
  \emph{Proceedings of the International Conference on Machine Learning
  (ICML)}, ser. Proceedings of Machine Learning Research, vol. 119.\hskip 1em
  plus 0.5em minus 0.4em\relax PMLR, 2020, pp. 5132--5143. [Online]. Available:
  \url{https://proceedings.mlr.press/v119/karimireddy20a.html}
\BIBentrySTDinterwordspacing

\bibitem{gu2017badnets}
\BIBentryALTinterwordspacing
T.~Gu, B.~Dolan-Gavitt, and S.~Garg, ``{BadNets}: Identifying vulnerabilities
  in the machine learning model supply chain,'' 2017. [Online]. Available:
  \url{https://arxiv.org/abs/1708.06733}
\BIBentrySTDinterwordspacing

\bibitem{liu2018trojannn}
\BIBentryALTinterwordspacing
Y.~Liu, S.~Ma, Y.~Aafer, W.-C. Lee, J.~Zhai, W.~Wang, and X.~Zhang, ``Trojaning
  attack on neural networks,'' in \emph{Proceedings of the Network and
  Distributed System Security Symposium (NDSS)}, 2018. [Online]. Available:
  \url{https://doi.org/10.14722/ndss.2018.23291}
\BIBentrySTDinterwordspacing

\bibitem{nguyen2020inputaware}
\BIBentryALTinterwordspacing
T.~A. Nguyen and A.~Tran, ``Input-aware dynamic backdoor attack,'' in
  \emph{Advances in Neural Information Processing Systems (NeurIPS)}, 2020.
  [Online]. Available:
  \url{https://proceedings.neurips.cc/paper/2020/hash/234e691320c0ad5b45ee3c96d0d7b8f8-Abstract.html}
\BIBentrySTDinterwordspacing

\bibitem{wang2020attackoftails}
\BIBentryALTinterwordspacing
H.~Wang, K.~Sreenivasan, S.~Rajput, H.~Vishwakarma, S.~Agarwal, J.~yong Sohn,
  K.~Lee, and D.~Papailiopoulos, ``Attack of the tails: Yes, you really can
  backdoor federated learning,'' in \emph{Advances in Neural Information
  Processing Systems (NeurIPS)}, 2020. [Online]. Available:
  \url{https://proceedings.neurips.cc/paper_files/paper/2020/hash/b8ffa41d4e492f0fad2f13e29e1762eb-Abstract.html}
\BIBentrySTDinterwordspacing

\bibitem{dai2023chameleon}
\BIBentryALTinterwordspacing
Y.~Dai and S.~Li, ``Chameleon: Adapting to peer images for planting durable
  backdoors in federated learning,'' in \emph{Proceedings of the International
  Conference on Machine Learning (ICML)}, ser. Proceedings of Machine Learning
  Research, vol. 202.\hskip 1em plus 0.5em minus 0.4em\relax PMLR, 2023.
  [Online]. Available: \url{https://proceedings.mlr.press/v202/dai23a.html}
\BIBentrySTDinterwordspacing

\bibitem{yang2025lga}
\BIBentryALTinterwordspacing
Q.~Yang, P.~Yan, X.~Wu, J.~Zhang, T.~Song, Y.~Hua, H.~Wang, L.~Wang, and
  H.~Guan, ``Stealthy backdoor attack in federated learning via adaptive
  layer-wise gradient alignment,'' in \emph{Proceedings of the IEEE/CVF
  International Conference on Computer Vision (ICCV)}, 2025, pp.
  29\,163--29\,172. [Online]. Available:
  \url{https://openaccess.thecvf.com/content/ICCV2025/html/Yang_Stealthy_Backdoor_Attack_in_Federated_Learning_via_Adaptive_Layer-wise_Gradient_ICCV_2025_paper.html}
\BIBentrySTDinterwordspacing

\bibitem{lyu2023cerberus}
\BIBentryALTinterwordspacing
X.~Lyu, Y.~Han, W.~Wang, J.~Liu, B.~Wang, J.~Liu, and X.~Zhang, ``Poisoning
  with cerberus: Stealthy and colluded backdoor attack against federated
  learning,'' in \emph{Proceedings of the AAAI Conference on Artificial
  Intelligence}, 2023. [Online]. Available:
  \url{https://ojs.aaai.org/index.php/AAAI/article/view/26083}
\BIBentrySTDinterwordspacing

\bibitem{liu2024fcba}
\BIBentryALTinterwordspacing
T.~Liu, Y.~Zhang, Z.~Feng, Z.~Yang, C.~Xu, D.~Man, and W.~Yang, ``Beyond
  traditional threats: A persistent backdoor attack on federated learning,'' in
  \emph{Proceedings of the AAAI Conference on Artificial Intelligence}, 2024,
  pp. 21\,359--21\,367. [Online]. Available:
  \url{https://ojs.aaai.org/index.php/AAAI/article/view/30131}
\BIBentrySTDinterwordspacing

\bibitem{he2026batman}
\BIBentryALTinterwordspacing
W.~He, W.~Huang, Y.~Fang, W.~Qu, J.~Zhang, and M.~Ye, ``Batman: Benign
  knowledge alignment through malicious null space in federated backdoor
  attack,'' in \emph{Proceedings of the IEEE/CVF Conference on Computer Vision
  and Pattern Recognition (CVPR)}, 2026, pp. 13\,316--13\,325. [Online].
  Available:
  \url{https://openaccess.thecvf.com/content/CVPR2026/html/He_Batman_Benign_Knowledge_Alignment_Through_Malicious_Null_Space_in_Federated_CVPR_2026_paper.html}
\BIBentrySTDinterwordspacing

\bibitem{ye2026lessismore}
\BIBentryALTinterwordspacing
P.~Ye, Y.~Li, K.~He, H.~Wang, R.~Du, and W.~Wang, ``Less is more: Persistent
  low-frequency backdoor injection in federated learning,'' in
  \emph{Proceedings of the IEEE International Conference on Computer
  Communications (INFOCOM)}, 2026. [Online]. Available:
  \url{https://www.cse.ust.hk/~weiwa/papers/ye-infocom26.pdf}
\BIBentrySTDinterwordspacing

\bibitem{lyu2025coba}
\BIBentryALTinterwordspacing
X.~Lyu, Y.~Han, W.~Wang, J.~Liu, B.~Wang, K.~Chen, Y.~Li, J.~Liu, and X.~Zhang,
  ``{CoBA}: Collusive backdoor attacks with optimized trigger to federated
  learning,'' \emph{IEEE Transactions on Dependable and Secure Computing},
  vol.~22, no.~2, pp. 1506--1518, 2025. [Online]. Available:
  \url{https://dblp.org/rec/journals/tdsc/LyuHWLWCLLZ25}
\BIBentrySTDinterwordspacing

\bibitem{li2025mirage}
\BIBentryALTinterwordspacing
Y.~Li, Y.~Zhao, C.~Zhu, and J.~Zhang, ``Infighting in the dark: Multi-label
  backdoor attack in federated learning,'' in \emph{Proceedings of the IEEE/CVF
  Conference on Computer Vision and Pattern Recognition (CVPR)}, 2025, pp.
  25\,770--25\,779. [Online]. Available:
  \url{https://openaccess.thecvf.com/content/CVPR2025/papers/Li_Infighting_in_the_Dark_Multi-Label_Backdoor_Attack_in_Federated_Learning_CVPR_2025_paper.pdf}
\BIBentrySTDinterwordspacing

\bibitem{feng2025sadba}
\BIBentryALTinterwordspacing
J.~Feng, Y.~Lai, H.~Sun, and B.~Ren, ``{SADBA}: Self-adaptive distributed
  backdoor attack against federated learning,'' in \emph{Proceedings of the
  AAAI Conference on Artificial Intelligence}, 2025, pp. 16\,568--16\,576.
  [Online]. Available:
  \url{https://ojs.aaai.org/index.php/AAAI/article/view/33820}
\BIBentrySTDinterwordspacing

\bibitem{yin2018byzantine}
\BIBentryALTinterwordspacing
D.~Yin, Y.~Chen, R.~Kannan, and P.~Bartlett, ``Byzantine-robust distributed
  learning: Towards optimal statistical rates,'' in \emph{Proceedings of the
  International Conference on Machine Learning (ICML)}, ser. Proceedings of
  Machine Learning Research, vol.~80.\hskip 1em plus 0.5em minus 0.4em\relax
  PMLR, 2018, pp. 5650--5659. [Online]. Available:
  \url{https://proceedings.mlr.press/v80/yin18a.html}
\BIBentrySTDinterwordspacing

\bibitem{blanchard2017krum}
\BIBentryALTinterwordspacing
P.~Blanchard, E.~M.~E. Mhamdi, R.~Guerraoui, and J.~Stainer, ``Machine learning
  with adversaries: Byzantine tolerant gradient descent,'' in \emph{Advances in
  Neural Information Processing Systems (NeurIPS)}, 2017, pp. 119--129.
  [Online]. Available: \url{https://dl.acm.org/doi/abs/10.5555/3294771.3294783}
\BIBentrySTDinterwordspacing

\bibitem{pillutla2022rfa}
\BIBentryALTinterwordspacing
K.~Pillutla, S.~M. Kakade, and Z.~Harchaoui, ``Robust aggregation for federated
  learning,'' \emph{IEEE Transactions on Signal Processing}, vol.~70, pp.
  1142--1154, 2022. [Online]. Available:
  \url{https://ieeexplore.ieee.org/document/9714365}
\BIBentrySTDinterwordspacing

\bibitem{fung2020foolsgold}
\BIBentryALTinterwordspacing
C.~Fung, C.~J.~M. Yoon, and I.~Beschastnikh, ``The limitations of federated
  learning in sybil settings,'' in \emph{Proceedings of the International
  Symposium on Research in Attacks, Intrusions and Defenses (RAID)}.\hskip 1em
  plus 0.5em minus 0.4em\relax USENIX Association, 2020, pp. 301--316.
  [Online]. Available:
  \url{https://www.usenix.org/conference/raid2020/presentation/fung}
\BIBentrySTDinterwordspacing

\bibitem{cao2021fltrust}
\BIBentryALTinterwordspacing
X.~Cao, M.~Fang, J.~Liu, and N.~Z. Gong, ``{FLTrust}: Byzantine-robust
  federated learning via trust bootstrapping,'' in \emph{Proceedings of the
  Network and Distributed System Security Symposium (NDSS)}, 2021. [Online].
  Available: \url{https://doi.org/10.14722/ndss.2021.24434}
\BIBentrySTDinterwordspacing

\bibitem{abadi2016dp}
\BIBentryALTinterwordspacing
M.~Abadi, A.~Chu, I.~Goodfellow, H.~B. McMahan, I.~Mironov, K.~Talwar, and
  L.~Zhang, ``Deep learning with differential privacy,'' in \emph{Proceedings
  of the ACM SIGSAC Conference on Computer and Communications Security
  (CCS)}.\hskip 1em plus 0.5em minus 0.4em\relax ACM, 2016, pp. 308--318.
  [Online]. Available: \url{https://doi.org/10.1145/2976749.2978318}
\BIBentrySTDinterwordspacing

\bibitem{xie2021crfl}
\BIBentryALTinterwordspacing
C.~Xie, M.~Chen, P.-Y. Chen, and B.~Li, ``{CRFL}: Certifiably robust federated
  learning against backdoor attacks,'' in \emph{Proceedings of the
  International Conference on Machine Learning (ICML)}, ser. Proceedings of
  Machine Learning Research, vol. 139.\hskip 1em plus 0.5em minus 0.4em\relax
  PMLR, 2021. [Online]. Available:
  \url{https://proceedings.mlr.press/v139/xie21a.html}
\BIBentrySTDinterwordspacing

\bibitem{ozdayi2021robustlr}
\BIBentryALTinterwordspacing
M.~S. Ozdayi, M.~Kantarcioglu, and Y.~R. Gel, ``Defending against backdoors in
  federated learning with robust learning rate,'' in \emph{Proceedings of the
  AAAI Conference on Artificial Intelligence}, vol.~35, no.~10, 2021, pp.
  9268--9276. [Online]. Available:
  \url{https://ojs.aaai.org/index.php/AAAI/article/view/17118}
\BIBentrySTDinterwordspacing

\bibitem{nguyen2022flame}
\BIBentryALTinterwordspacing
T.~D. Nguyen, P.~Rieger, H.~Fereidooni, S.~Marchal, M.~Miettinen,
  A.~Mirhoseini, A.-R. Sadeghi, T.~Schneider, and S.~Ma, ``{FLAME}: Taming
  backdoors in federated learning,'' in \emph{USENIX Security Symposium}.\hskip
  1em plus 0.5em minus 0.4em\relax USENIX Association, 2022, pp. 1415--1432.
  [Online]. Available:
  \url{https://www.usenix.org/conference/usenixsecurity22/presentation/nguyen}
\BIBentrySTDinterwordspacing

\bibitem{rieger2022deepsight}
\BIBentryALTinterwordspacing
P.~Rieger, T.~D. Nguyen, M.~Miettinen, and A.-R. Sadeghi, ``{DeepSight}:
  Mitigating backdoor attacks in federated learning through deep model
  inspection,'' in \emph{Proceedings of the Network and Distributed System
  Security Symposium (NDSS)}, 2022. [Online]. Available:
  \url{https://www.ndss-symposium.org/ndss-paper/auto-draft-205/}
\BIBentrySTDinterwordspacing

\bibitem{zhang2023flip}
\BIBentryALTinterwordspacing
K.~Zhang, G.~Tao, Q.~Xu, S.~Cheng, S.~An, Y.~Liu, S.~Feng, G.~Shen, P.-Y. Chen,
  S.~Ma, and X.~Zhang, ``{FLIP}: A provable defense framework for backdoor
  mitigation in federated learning,'' in \emph{Proceedings of the International
  Conference on Learning Representations (ICLR)}, 2023. [Online]. Available:
  \url{https://openreview.net/forum?id=Xo2E217_M4n}
\BIBentrySTDinterwordspacing

\bibitem{huang2023lockdown}
\BIBentryALTinterwordspacing
T.~Huang, S.~Hu, K.-H. Chow, F.~Ilhan, S.~F. Tekin, and L.~Liu, ``Lockdown:
  Backdoor defense for federated learning with isolated subspace training,'' in
  \emph{Advances in Neural Information Processing Systems (NeurIPS)}, 2023.
  [Online]. Available:
  \url{https://proceedings.neurips.cc/paper_files/paper/2023/hash/2376f25ef1725a9e3516ee3c86a59f46-Abstract-Conference.html}
\BIBentrySTDinterwordspacing

\bibitem{li2024backdoorindicator}
\BIBentryALTinterwordspacing
S.~Li and Y.~Dai, ``{BackdoorIndicator}: Leveraging {OOD} data for proactive
  backdoor detection in federated learning,'' in \emph{USENIX Security
  Symposium}.\hskip 1em plus 0.5em minus 0.4em\relax USENIX Association, 2024,
  pp. 4193--4210. [Online]. Available:
  \url{https://www.usenix.org/conference/usenixsecurity24/presentation/li-songze}
\BIBentrySTDinterwordspacing

\bibitem{kabir2024flshield}
\BIBentryALTinterwordspacing
E.~Kabir, Z.~Song, M.~R.~U. Rashid, and S.~Mehnaz, ``{FLShield}: A validation
  based federated learning framework to defend against poisoning attacks,'' in
  \emph{Proceedings of the IEEE Symposium on Security and Privacy (S\&P)},
  2024, pp. 2572--2590. [Online]. Available:
  \url{https://doi.org/10.1109/SP54263.2024.00141}
\BIBentrySTDinterwordspacing

\bibitem{xu2025alignins}
\BIBentryALTinterwordspacing
J.~Xu, Z.~Zhang, and R.~Hu, ``Detecting backdoor attacks in federated learning
  via direction alignment inspection,'' in \emph{Proceedings of the IEEE/CVF
  Conference on Computer Vision and Pattern Recognition (CVPR)}, 2025, pp.
  20\,654--20\,664. [Online]. Available:
  \url{https://openaccess.thecvf.com/content/CVPR2025/papers/Xu_Detecting_Backdoor_Attacks_in_Federated_Learning_via_Direction_Alignment_Inspection_CVPR_2025_paper.pdf}
\BIBentrySTDinterwordspacing

\bibitem{wan2025mars}
\BIBentryALTinterwordspacing
W.~Wan, Y.~Ning, Z.~Huang, C.~Hong, S.~Hu, Z.~Zhou, Y.~Zhang, T.~Zhu, W.~Zhou,
  and L.~Y. Zhang, ``{MARS}: A malignity-aware backdoor defense in federated
  learning,'' in \emph{Advances in Neural Information Processing Systems
  (NeurIPS)}, 2025. [Online]. Available:
  \url{https://openreview.net/forum?id=3kmbucBZPA}
\BIBentrySTDinterwordspacing

\bibitem{yu2026guardfl}
\BIBentryALTinterwordspacing
H.~Yu, C.~Ma, M.~Liu, T.~Du, M.~Ding, T.~Xiang, S.~Ji, and X.~Liu,
  ``{G2uardFL}: Safeguarding federated learning against backdoor attacks via
  attributed client graph clustering,'' \emph{IEEE Transactions on Information
  Forensics and Security}, vol.~21, pp. 516--531, 2026. [Online]. Available:
  \url{https://doi.org/10.1109/TIFS.2025.3639985}
\BIBentrySTDinterwordspacing

\bibitem{krizhevsky2009cifar}
\BIBentryALTinterwordspacing
A.~Krizhevsky and G.~Hinton, ``Learning multiple layers of features from tiny
  images,'' University of Toronto, Toronto, ON, Canada, Tech. Rep., 2009.
  [Online]. Available:
  \url{https://www.cs.toronto.edu/~kriz/learning-features-2009-TR.pdf}
\BIBentrySTDinterwordspacing

\bibitem{le2015tinyimagenet}
\BIBentryALTinterwordspacing
Y.~Le and X.~Yang, ``{Tiny ImageNet} visual recognition challenge,'' \emph{CS
  231N}, vol.~7, no.~7, p.~3, 2015. [Online]. Available:
  \url{https://cs231n.stanford.edu/reports/2015/pdfs/yle_project.pdf}
\BIBentrySTDinterwordspacing

\bibitem{he2016resnet}
\BIBentryALTinterwordspacing
K.~He, X.~Zhang, S.~Ren, and J.~Sun, ``Deep residual learning for image
  recognition,'' in \emph{Proceedings of the IEEE/CVF Conference on Computer
  Vision and Pattern Recognition (CVPR)}, 2016, pp. 770--778. [Online].
  Available: \url{https://doi.org/10.1109/CVPR.2016.90}
\BIBentrySTDinterwordspacing

\end{thebibliography}

\end{document}